\pdfoutput=1

\documentclass[11pt]{article}
\usepackage[]{EMNLP2023}
\usepackage{times}
\usepackage{latexsym}
\usepackage{CJKutf8}

\usepackage{amsmath}
\usepackage{graphicx}
\graphicspath{{images/}}
\usepackage{blindtext}
\usepackage{amssymb}
\usepackage{multirow}
\usepackage{array}
\usepackage{booktabs}
\usepackage{tabularx,ragged2e}
\usepackage{import}
\usepackage[T1]{fontenc}
\usepackage[utf8]{inputenc}
\usepackage{microtype}
\usepackage{inconsolata}
\usepackage{blindtext}

\title{Diversity Enhanced Narrative Question Generation for StoryBooks}
\author{Hokeun Yoon \\
  Sungkyunkwan University\\
  Suwon, South Korea \\
  \texttt{hkyoon95@g.skku.edu} \\
  \And
  JinYeong Bak \\
  Sungkyunkwan University\\
  Suwon, South Korea \\
  \texttt{jy.bak@skku.edu} \\
  }
\date{Dec 2023}

\begin{document}
\maketitle
\begin{abstract}
Question generation (QG) from a given context can enhance comprehension, engagement, assessment, and overall efficacy in learning or conversational environments. Despite recent advancements in QG, the challenge of enhancing or measuring the diversity of generated questions often remains unaddressed. In this paper, we introduce a multi-question generation model (mQG), which is capable of generating multiple, diverse, and answerable questions by focusing on context and questions. To validate the answerability of the generated questions, we employ a SQuAD2.0 fine-tuned question answering model, classifying the questions as answerable or not. We train and evaluate mQG on the FairytaleQA dataset, a well-structured QA dataset based on storybooks, with narrative questions. We further apply a zero-shot adaptation on the TellMeWhy and SQuAD1.1 datasets. mQG shows promising results across various evaluation metrics, among strong baselines.\footnote{Code: \url{https://github.com/hkyoon95/mQG}}

\end{abstract}

\section{Introduction}
\label{sec:intro}


Question generation (QG), focusing on the questions derived from specific text passages or documents, plays an integral role in a wide array of domains. It improves question answering (QA) systems \citep{sultan-etal-2020-importance}, enriches educational experiences \citep{yao-etal-2022-ais}, and enhances the engagement factor in chatbots \citep{laban-etal-2020-whats}. The effectiveness of QG tasks can be significantly improved by generating multiple questions, ensuring a broader, more comprehensive exploration of the content.

The importance of generating and evaluating multiple questions becomes evident when we examine the creation process of QA datasets \citep{richardson-etal-2013-mctest, rajpurkar-etal-2016-squad, xu-etal-2022-fantastic}. Traditional QA dataset creation typically involves instructing annotators to create a pre-determined number of questions for a given context. Recent QG research \citep{wang-etal-2020-pathqg, yao-etal-2022-ais}, however, tends to rely on automatic evaluation of semantic similarity with golden questions, often overlooking the potential for diverse aspects of questions. When generating multiple questions, diversity is a crucial aspect to consider. The diversity of questions can span several dimensions, including varied aspects of the context, different answer types, and different phrasings for essentially the same question \citep{Karttunen1977-KARSAS}. This diversity allows for a more comprehensive exploration of the context. The diversity of questions can be broadly measured based on the type of answers they require; explicit questions with answers that can be explicitly found in the reading materials, and implicit questions with answers that require deductive reasoning. The crafting of multiple questions, bearing in mind both diversity and alignment with reading materials, poses a cognitively demanding and time-consuming task for humans.






One significant application of generating diverse and multiple questions is education. It has been observed that children can develop better reading comprehension skills at an early age by creating narrative questions themselves and being asked comprehension-related questions about storybooks \citep{FrancisDavidJ2005DAtA, Janssen_Braaksma_Couzijn_2009}. Reading comprehension is an essential skill that requires learners to combine knowledge and reason about relations, entities, and events across a given context \citep{doi:10.1080/10888438.2017.1291643, MohseniTakaloo}.
Consequently, a system that can generate diverse and multiple narrative questions can serve as a valuable enhancement to educational resources, aiding in student engagement and promoting a deeper understanding of study materials.

Recently, some researchers have attempted to generate multiple narrative questions. For educational applications, \citet{yao-etal-2022-ais} proposed to generate question-answer pairs with a three-step pipeline. As they use heuristic-generated answers to generate narrative questions most of their outcome is restricted to explicit questions. Also, \citet{zhao-etal-2022-educational} proposed to generate certain types of narrative questions and they tried to restrict the number of generated questions to a number of ground-truth questions, insisting that knowing question type distribution for each context is a sub-skill in education \citep{article}. We set these two approaches as our main baselines.

To address the above challenges, we introduce a multi-question generation model (mQG) that generates diverse and contextually relevant questions by referencing questions from the same context. mQG is trained with maximum question similarity loss $L_{MQS}$, which is designed to make the representation of reference questions and the representation of a target question similar. Moreover, mQG employs a recursive generation framework, where previously generated questions are recursively fed back into the model as mQG is trained to output different questions from reference questions. Same as our two baselines, mQG is trained and evaluated on the FairytaleQA dataset, which focuses on narrative comprehension of storybooks. This dataset is designed to provide high-quality narrative QA pairs for students from kindergarten to eighth grade (ages 4 to 14), and labeled questions as explicit or implicit. We adopt Self-BLEU \citep{10.1145/3209978.3210080} to evaluate the diversity of generated questions. Beyond diversity, to consider generated questions relevant to the context, we demonstrate the answerability evaluation model to assess whether the generated questions are answerable. We also evaluate on TellMeWhy \citep{lal-etal-2021-tellmewhy} and SQuAD1.1 \citep{rajpurkar-etal-2016-squad} datasets with zero-shot adaptation to further analyze the performance of mQG in different settings. Differing from previous approaches, mQG successfully generates a substantial number of diverse and answerable narrative questions.

The main contributions of this paper are summarized as follows.
\begin{itemize}
\item We expand the scope of the question generation task by generating a comprehensive set of questions, regardless of our knowledge of the answers, and subsequently categorize them into answerable and non-answerable questions.

\item We introduce mQG, a novel question generation model that is trained using the maximum question similarity loss $L_{MQS}$ and employs a recursive referencing process for generating a wide array of questions while preserving semantic correctness.

\item We introduce an answerability evaluation model capable of classifying questions as implicit, explicit, or unanswerable.

\end{itemize}


\section{Related Work}
\label{sec:related}
\subsection{Question Generation}

Based on given contents, question generation aims to generate natural language questions, where the generated questions are able to be addressed with the given contents. After neural approaches took over a large proportion in QG \citep{DBLP:journals/corr/YuanWGSBSZT17, DBLP:journals/corr/ZhouYWTBZ17}, QG can largely be separated by target answer aspect into answer-aware QG and answer-unaware QG. Answer-aware QG, as its name implies, provides an answer to a model and prompts it to generate questions based on those answers. On the other hand, answer-unaware QG mainly focuses on the context to formulate questions. The introduction of pre-trained Language Models (LMs) further accelerated advancements in QG, and many works have demonstrated significant improvement in the answer-aware QG task and presented promising possibilities for QG \citep{zhang-bansal-2019-addressing, DBLP:journals/corr/abs-1905-03197, DBLP:journals/corr/abs-2001-04063}. This approach inherently favors explicit questions, which can be directly answered with the provided context. In answer-unaware QG, only a handful of studies have been conducted, primarily focusing on strategies such as sentence selection from a paragraph \citep{du-cardie-2017-identifying}, employing transformer architectures with out-of-vocabulary methods \citep{scialom-etal-2019-self}, and generating questions based on silver summaries \citep{zhao-etal-2022-educational}. In this paper, we utilize answer-unaware question generation, giving consideration to both the diversity and quality of explicit and implicit questions.

\begin{figure*}[ht]
\centerline
{\includegraphics[height=6.5cm]{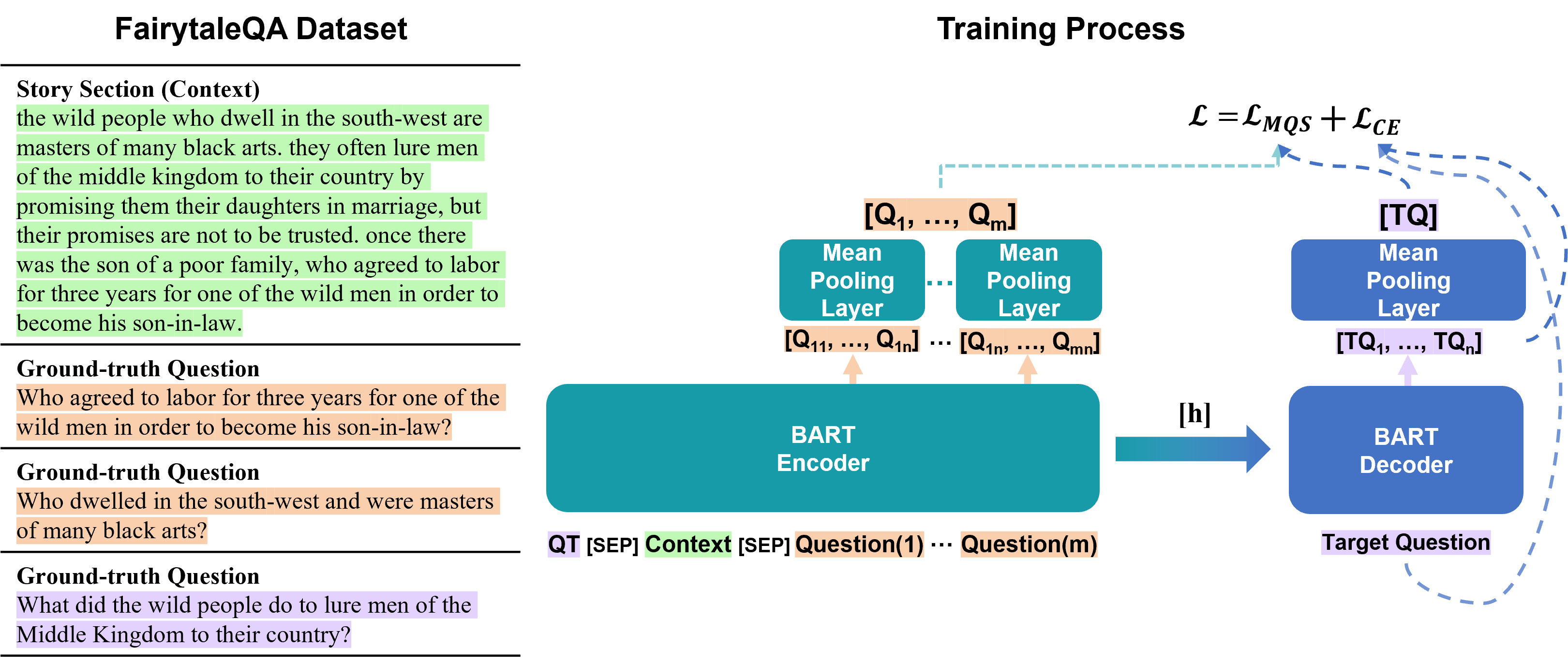}}
\caption{Overview of the training process of mQG. 
$Question(1)$ to $Question(m)$ refer to ground-truth questions from the same context (orange), without a ground-truth question (purple) input to BART Decoder. 
$QT$ and $[h]$ denote the wh-word corresponding to the target question and overall encoder representation.
}
\label{fig:mQG}
\end{figure*}

\subsection{Diversity}

In natural language generation (NLG), generating outputs that are not only correct but also diverse is essential. In the decoding aspect, diversity has been researched in areas such as top-k sampling \citep{fan-etal-2018-hierarchical}, and nucleus sampling \citep{Holtzman2020The}. These decoding methods tried to sample tokens from less likely vocabularies. Certain studies have focused on training models to yield more diverse outputs \citep{Welleck2020Neural, yao-etal-2022-ais}, and on leveraging the combination of contrastive training and generation \citep{su2022a}. Recently, \citet{sultan-etal-2020-importance} evaluated the importance of diversity in QG, insisting that diverse and accurate questions yield better QA results. Additionally, some researchers explored diversity in QG based on relevant topic \citep{hu2018topicbased}, content selectors with question type modeling \citep{wang-etal-2020-diversify}, control of question type \citep{DBLP:journals/corr/abs-2107-00152}, and difficulty level \citep{cheng-etal-2021-guiding}. While these studies have addressed various aspects of diversity in QG, there is still considerable room for further research in this area. In this paper, we consider diversity a significant challenge in the question generation task and propose a model that can generate a wide range of answerable questions.

\section{Method}
\label{sec:method}
\graphicspath{{\subfix{../images/}}}

In this section, we formalize the multi-question generation task and introduce our mQG. We first formulate our task and then explain how our model's training process incorporates a maximum question similarity loss $\mathcal{L}_{MQS}$. Finally, we provide a detailed outline of our recursive generation framework.

\subsection{Task Formulation}
 \label{sec:task}
 The QG task in this paper aims to generate each question using a given context, question type, and the history of questions generated from the same context with the same question type.
 We use seven wh-words (what, when, where, which, who, why, how) as question types. 
 Mathematically, given the context $C$, question type $QT$, and history of generated questions $H_i = (GQ_1, GQ_2, ..., GQ_{i-1})$, this task can be defined as generating a question, $\hat{GQ}$, where:
\begin{equation}
\hat{GQ} = \operatorname{{argmax}_{GQ_i}}(Prob(GQ_i | QT, C, H_i))
\end{equation}
For the training process, we extract wh-words from each question by applying part-of-speech tagging with the Spacy\footnote{\url{https://spacy.io/}} English Model. Due to the absence of a history of generated questions and an insufficient number of questions per context per question type in the FairytaleQA dataset, we utilize ground-truth questions that only share the context as the history of questions within the training process. 

\subsection{Diversity Enhanced Training}
\label{sec:qcl}

mQG is built upon BART \citep{lewis-etal-2020-bart}, which has demonstrated remarkable performance in various natural language processing tasks. The primary pre-training objective of BART is to reconstruct the masked input based on unmasked input. To further leverage the capabilities of the pre-trained BART, we introduce a maximum question similarity loss $\mathcal{L}_{MQS}$. This loss is designed to promote similar representations for different questions from the encoder and decoder.

As shown in Figure \ref{fig:mQG}, the encoder takes in three inputs: the question type, which signifies the type of question to be generated; the context, which provides the necessary information for question generation; and ground-truth questions from the same context, serving as reference questions. These three inputs are concatenated, with a \texttt{[SEP]} token inserted between them.  The encoder processes the input sequence and produces its corresponding representations. Subsequently, the decoder generates the representation for the target question. To calculate the maximum question similarity loss $\mathcal{L}_{MQS}$, we use mean pooling layers to convert question representations into sentence-level representations. The maximum question similarity loss $\mathcal{L}_{MQS}$ is calculated between the sentence-level representation of the reference questions and the sentence-level representation of a generated question. By encouraging the representation of different questions to be similar, we promote the generation of diverse questions that differ from reference questions.

Given a set of reference questions sentence-level representation as $Q = \{{Q_1, ..., Q_m}\}$ and a sentence-level representation of the target question as $TQ$, the maximum question similarity loss $\mathcal{L}_{MQS}$ is computed as follows:
\begin{equation}
\begin{aligned}
\mathcal{L}_{MQS} = & \frac{1}{m} \sum_{i=1}^{{m}} \max(0, 1-s(Q_i,TQ))
\end{aligned}
\end{equation}
where $s(Q_i,TQ)$ is a cosine similarity calculation between representations. By optimizing the model parameters to maximize the sentence-level similarity between these different representations, we guide mQG to generate diverse questions within the range of semantic correctness. This is achieved by ensuring that all the representations, which are the ground truth questions, are semantically correct. In doing so, we maintain a balance between diversity and accuracy in the generated questions. The overall training objective $\mathcal{L}$ is defined as 
\begin{equation}
\begin{aligned}
\mathcal{L} = \mathcal{L}_{CE} + \mathcal{L}_{MQS} 
\end{aligned}
\end{equation}
$\mathcal{L}_{CE}$ refers to the cross-entropy loss from a target question. As cross-entropy loss is calculated at the token level, the use of cross-entropy loss enhances mQG to generate syntactically correct questions.

\subsection{Recursive Generation Framework}

\begin{figure}[t]
\centering
{\includegraphics[scale=0.5]{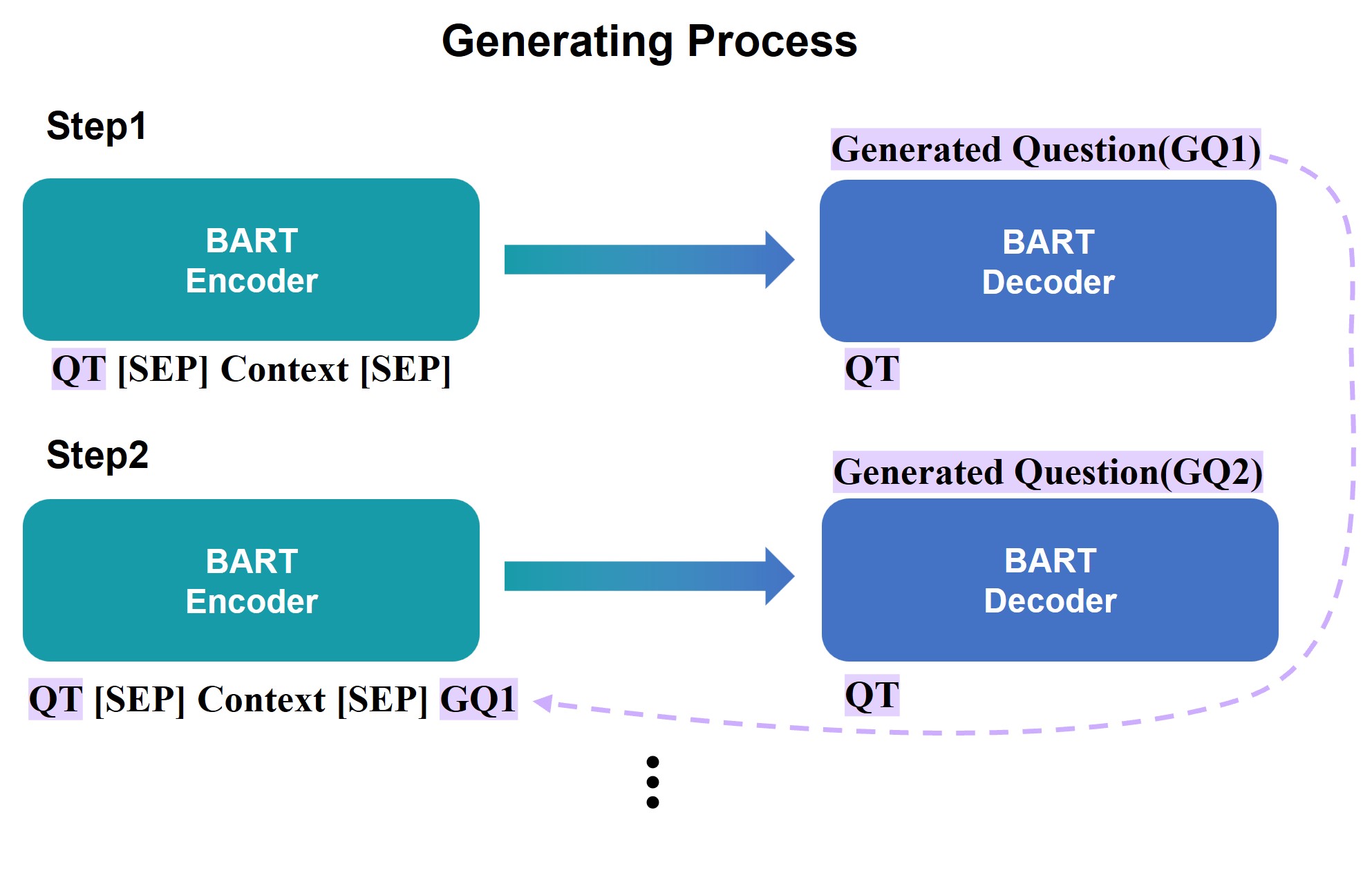}}
\caption{
The Recursive Generation Framework of mQG.
This framework involves an iterative process, using previously generated questions as input for subsequent steps, thereby creating a recursive cycle.
Each iteration maintains the use of the same question type.
}
\label{fig:mQG-gen}
\end{figure}

Figure \ref{fig:mQG-gen} illustrates the generation process of mQG. First, the encoder takes question type, and context as input. The decoder then generates a question based on the information provided by the encoder. For the subsequent generation steps, the previously generated questions are recursively fed back into the model. Specifically, the previous questions are concatenated with the same question type and context, separated by a \texttt{[SEP]} token. This concatenated sequence is then used as input for the next generation step. This recursive generation process continues until the desired number of questions per context per question type is achieved.

The use of this recursive generation process allows mQG to generate multiple questions while considering the previously generated questions. Following the training process of mQG, this generation process enables mQG to build upon its own previous outputs and generate different questions from previous outputs. We use beam search for the decoding method and return multiple sequences to exclude pre-generated questions. By leveraging a recursive framework, mQG demonstrates its capability to generate a variety of diverse questions that are contextually relevant and coherent.

\section{Experiments}
\label{sec:exp}
\subsection{Dataset}
\label{data}
\textbf{FairytaleQA \citep{xu-etal-2022-fantastic}}. We train mQG with the FairytaleQA dataset, which is constructed for educational purposes. Each book is split into sections and annotators were instructed to create on average 2-3 narrative question-answer pairs per section. All question-answer pairs are annotated based on seven question types that capture narrative elements/relations. Questions are labeled as explicit or implicit questions based on whether or not the answer source can be directly found in the context. The original FairytaleQA dataset is constructed in a train/validation/test set with 232/23/23 books and 8,548/1,025/1,007 QA pairs. From the entire dataset, a small portion of questions (985 out of 10,580) spans multiple paragraphs. As mQG and baselines are fit for one paragraph we remove those questions. To cross-validate, we randomly shuffled the dataset and split it by books in train/validation/test set with roughly matching 80/10/10 (\%).

\subsection{Baselines}
In the experiments, we compare mQG with four baselines; an end-to-end model initialized with BART-large, and methods proposed in \citet{su2022a}, \citet{yao-etal-2022-ais}, \citet{zhao-etal-2022-educational} denoted as CB, QAG, and EQG. The last two baselines are designed for multiple question generation purposes.
\\
\\
\textbf{E2E.} As the FairytaleQA dataset consists of multiple questions in one context, we concat all questions and train the BART-large model to generate questions based on each context. To match the number of generated questions, we set the maximal target length to 280 tokens which roughly matches the number of generated questions setting of mQG. 
\\
\\
\textbf{CB (Contrastive Baseline).} We construct this baseline following the framework in \citet{su2022a}, which tackles the problem of diversity in open-ended text generation. This framework first trains the language model using contrastive loss and decodes it with a contrastive search method. Since the contrastive baseline is proven for diverse text generation we apply it to GPT2 (denoted as CB (GPT2)), and BART (denoted as CB (BART)) and set it as our baseline. During generation, the maximal target length is set to 280 tokens.
\\
\\
\textbf{QAG.} This baseline follows a question-answer generation architecture by \citet{yao-etal-2022-ais}. This architecture first generates answers based on a heuristic-based answer generation module, which generates multiple answers per context. With the generated answers, BART generates corresponding questions. And, to verify the quality of the generated questions, DistilBERT ranking module ranks each QA pair and chooses the top questions. As our task is to generate multiple questions, we denote architecture without a ranking module as QAG and the top 10 questions per context chosen by the ranking module as QAG (top 10).
\\
\\
\textbf{EQG.} EQG model \citep{zhao-etal-2022-educational} generates questions based on silver summaries. Silver summary is a method proposed by \citet{DBLP:journals/corr/abs-1809-02922}, which inserts answers into the semantic parsed questions with a rule-based method. EQG consists of three steps: 1) generate question type distribution for each context with BERT; 2) generate silver summary with BART, using question type, question type ordering from a question type distribution module, and context; 3) generate question based on silver summary, question type, and question ordering with BART. Without a question type distribution module, EQG is able to generate multiple questions. Since our approach is to generate multiple questions we set the EQG baseline without question type distribution module.

\subsection{Automatic Evaluation}
\label{sec:autoeval}

\subsubsection{Evaluation Metrics}
\label{sec:evalmet}
In evaluating question generation, both the quality and diversity of the generated questions are critical components. Thus, we evaluate each aspect with separate automatic evaluation metrics. We use Rouge-L score \citep{lin-2004-rouge}, BERTScore \citep{zhang2020bertscore}, and BLEURT \citep{sellam-etal-2020-bleurt} to measure the quality of generated questions. Similar to \citet{yao-etal-2022-ais}, for each ground-truth question, we find the highest semantic similarity score on generated questions from the same context than average overall semantic similarity scores. And, with multiple questions generated from the same context, we recognize the necessity to measure diversity automatically. For diversity measurement, we use Self-BLEU score \citep{10.1145/3209978.3210080} which was introduced to evaluate just a variety of sentences. The Self-BLEU score, which uses each generated sentence as a hypothesis and others as references, is employed to evaluate the diversity of questions generated from the same context. A lower Self-BLEU score represents greater diversity. All metrics ranges are between 0 to 1 except Rouge-L score (0 to 100).

\begin{table*}[ht]
\centering
\resizebox{0.57\textheight}{!}{\begin{tabular}{lrr rr rr rr cc cc}
\hline
\toprule
\multicolumn{13}{c}{\begin{tabular}[c]{@{}c@{}c@{}} \textbf{FairytaleQA} \\ \end{tabular}} \\
\midrule
\multirow{2}{*}{\textbf{Architecture}} & \multicolumn{2}{c}{\begin{tabular}[c]{@{}c@{}c@{}} \textbf{\# Generated} \\ \textbf{Questions} \\ \textbf{Per Section} \end{tabular}} & \multicolumn{2}{c}{\begin{tabular}[c]{@{}c@{}c@{}} \textbf{\# Answerable} \\ \textbf{Questions} \\ \textbf{Per Section} $\uparrow$ \\ \end{tabular}} & \multicolumn{2}{c}{\begin{tabular}[c]{@{}c@{}}\textbf{Rouge-L} \\ \textbf{F1} $\uparrow$ \\ \end{tabular}} & \multicolumn{2}{c}{\begin{tabular}[c]{@{}c@{}}\textbf{BERTScore} \\ \textbf{F1} $\uparrow$ \\ \end{tabular}}& \multicolumn{2}{c}{\textbf{BLEURT} $\uparrow$} & \multicolumn{2}{c}{\textbf{Self-BLEU} $\downarrow$}  \\ 
\cmidrule(lr){2-3} \cmidrule(lr){4-5} \cmidrule(lr){6-7} \cmidrule(lr){8-9} \cmidrule(lr){10-11} \cmidrule(lr){12-13}
& \multicolumn{1}{c}{M} & \multicolumn{1}{c}{SE} & \multicolumn{1}{c}{M} & \multicolumn{1}{c}{SE} & \multicolumn{1}{c}{M} & \multicolumn{1}{c}{SE} & \multicolumn{1}{c}{M} & \multicolumn{1}{c}{SE} & \multicolumn{1}{c}{M} & \multicolumn{1}{c}{SE} & \multicolumn{1}{c}{M} & \multicolumn{1}{c}{SE} \\
\midrule
E2E         & 1.58 & 0.07 & 1.45 & 0.12 & 36.05 & 0.35 & 0.8960 & 0.0062 & 0.4064 & 0.0104 & - & - \\
CB (BART)   & 1.60 & 0.03 & 1.49 & 0.04 & 36.89 & 0.68 & 0.9074 & 0.0017 & 0.4045 & 0.0072 & - & - \\
CB (GPT2)   & 3.28 & 0.43 & 0.96 & 0.56 & 26.47 & 1.27 & 0.8937 & 0.0020 & 0.3328 & 0.0077 & 0.8906 & 0.0120 \\
EQG         & 28.00 & 0.00 & 3.80 & 0.76 & 41.05 & 1.61 & 0.9136 & 0.0034 & 0.4293 & 0.0118 & 0.9864 & 0.0043 \\
QAG (top10)  & 9.95 & 0.14 & 6.57 & 0.39 & 45.44 & 0.81 & 0.9208 & 0.0006 & 0.4444 & 0.0076 & 0.7608 & 0.0078 \\
QAG         & 26.97 & 0.50 & 15.95 & 1.24 & 53.77 & 1.03 & 0.9323 & 0.0009 & 0.5140 & 0.0115 & 0.8874 & 0.0030 \\
mQG    & 28.00 & 0.00 & \textbf{23.08} & 0.36 & \textbf{58.90} & 0.37 & \textbf{0.9394} & 0.0005 & \textbf{0.5698} & 0.0033 & \textbf{0.6389} & 0.0079 \\
\bottomrule
\hline
\end{tabular}}
\caption{Three cross-validation results on the FairytaleQA dataset. 
\# Answerable Questions Per Section is based on the answerability evaluation model, as described in section \ref{sec:evalmet}.  
$\uparrow$ means higher is better, $\downarrow$ means lower is better. 
Due to a low number of questions, Self-BLEU which cannot be measured is marked with a hyphen. M, SE denotes mean and standard error.
mQG generates the highest number of answerable questions with greater diversity.
}
\label{table:main}
\end{table*}

\subsubsection{Answerability Evaluation Model}

In order to evaluate whether the generated questions correspond to the context, we leverage SQuAD2.0 dataset \citep{rajpurkar-etal-2018-know} to build an evaluation model. SQuAD2.0 is a question-answering dataset with 100K answerable questions and 50K unanswerable questions. This dataset is used to enhance the evaluation model by classifying whether the questions are answerable or not. We use DeBERTa-base \citep{he2021deberta} as the backbone model.

To achieve our goal, we train the evaluation model on the QA task following implementation in \citet{devlin-etal-2019-bert}. We construct two dense layers above the encoder; one for the answer start position and the other for the answer end position. And, as unanswerable questions and implicit questions do not have an answer span, for these questions \texttt{[CLS]} token is assigned as the answer start position and the answer end position. For implicit questions in the FairytaleQA dataset, we add a special token \texttt{[IMP]} and assign it as an answer start span and answer end span. First, we train the evaluation model with the SQuAD2.0 dataset on the QA task. For the second step, we train the evaluation model again with the FairytaleQA dataset. By utilizing a two-step training, the evaluation model is able to classify generated questions as explicit, implicit, or unanswerable. The number of answerable questions per section in Table \ref{table:main} are based on classified results by the evaluation model. If the evaluation model classifies generated questions as implicit or explicit, then we count them as answerable. 
(Answerability evaluation model details are given in Appendix \ref{sec:fa_evalationmodel}.)

\subsubsection{Results}
\label{sec:result}

Table \ref{table:main} presents evaluation results on the FairytaleQA test set. `\# Generated Questions Per Section' refers to the number of questions generated for each section. In `\# Answerable Questions Per Section', as duplicate questions within the same context are not needed, we leave only one question from duplicate questions. Even though mQG is able to generate multiple questions within the maximum token length of BART, we roughly match the number of questions to QAG for fair comparison in Rouge-L F1, setting mQG to generate 4 questions per section per question type, totaling 28 questions per section. The same setting is applied to EQG, as EQG does not have limitations in generating multiple questions. 

General baselines (E2E and CB) that generate multiple questions in one iteration show significant underperformance in the Rouge-L F1 score and in the number of generated questions, compared to strong baselines (QAG and EQG), and the mQG. This indicates that to generate multiple questions, a specific model is needed. Across all evaluation metrics, mQG consistently outperforms the baselines.

\subsection{Human Evaluation}
We evaluate the diversity and quality of generated questions on the FairytaleQA dataset with human judges. We hire five annotators, proficient in English as their first foreign language, to further evaluate the diversity and quality of the generated questions. We follow the human evaluation procedure described by \citet{DBLP:journals/corr/abs-2107-00152} and compare mQG, with two robust baselines, EQG and QAG.
\\
\\
\textbf{Question Diversity.} In the question diversity study, we randomly sample 5 books from the original test set; and for each book, we randomly sample 8 sections, totaling 40 sections. For each section, we randomly sample three questions as a question set from each model, and provide only the question sets for annotation. For each question set, the annotators rank the three models on a scale of 1 (highest) to 3 (lowest) based on three dimensions of diversity: type–whether the three selected questions have different question types; syntax–whether the three selected questions use different syntax; and content–whether the three selected questions need to be addressed with diverse answers.

As shown in Table \ref{table:he1}, on all dimensions, human annotators rate mQG as generating the most diverse questions compared to the other models, with each question requiring a different answer.
\\
\\
\textbf{Question Quality.} In the question quality study, we again randomly sample 5 books from the original test set. For each book, we select a random sample of 8 sections. Each section contains four questions, each randomly sampled from three models and ground-truth, totaling 160 questions. Two dimensions are rated from 1 (worst) to 5 (best): appropriateness–whether the question is semantically correct; answerability–whether the question can be addressed by a given section.

As shown in Table \ref{table:he2}, all models, when compared to the ground-truth, generate semantically correct questions. Given that mQG can generate a broad diversity of questions, these results confirm that mQG fulfills our goal of generating multiple questions while maintaining semantic correctness and relevance to the context.

\begin{table}[t]
\centering
\resizebox{\columnwidth}{!}{\begin{tabular}{lccc}
\hline
\toprule
\multirow{1}{*}{\textbf{Architecture}} & \multicolumn{1}{c}{\textbf{Type (\%)}} &  \multicolumn{1}{c}{\textbf{Syntax (\%)}} & \multicolumn{1}{c}{\textbf{Content (\%)}} \\
\midrule
EQG & 22.0 & 18.0 & 23.5 \\
QAG & 33.0 & 22.0 & 34.5 \\
mQG & \textbf{77.0} & \textbf{70.5} & \textbf{60.0}\\
\bottomrule
\hline
\end{tabular}}
\caption{
Human evaluation on diversity. The percentage of samples ranked first among other models.
Krippendorf's alphas are 0.69, 0.51, and 0.38 for the three dimensions. 
Ties are allowed.
mQG demonstrates the most diversity in all dimensions.
}
\label{table:he1}
\end{table}

\begin{table}[t]
\centering
\resizebox{0.55\columnwidth}{!}{\begin{tabular}{lccc}
\hline
\toprule
\multirow{1}{*}{\textbf{Architecture}} & \multicolumn{1}{c}{\textbf{Appro.}} &  \multicolumn{1}{c}{\textbf{Ans.}} \\
\midrule
EQG & \textbf{4.85} & 4.46 \\
QAG & 4.60 & 4.43 \\
mQG & 4.79 & 4.47 \\
\hline
Ground-truth & 4.71 & \textbf{4.76} \\
\bottomrule
\hline
    \end{tabular}}
\caption{
 Human evaluation on appropriateness (Appro.) and answerability (Ans.).
 The Krippendorf's alphas are 0.14 and 0.27 for the two dimensions. Ties are allowed.
 In all models, not much difference is observed compared to ground truth questions.
 }
\label{table:he2}
\end{table}

\subsection{Zero-shot Performance Evaluation}
We conduct a zero-shot evaluation on two distinct datasets, to test mQG more in various real-world scenarios, where contexts and desired questions can differ. Zero-shot evaluation is essential for assessing model performance as it illuminates the model's ability to generalize beyond the specific examples it was trained on.

\subsubsection{Dataset}
\textbf{TellMeWhy \citep{lal-etal-2021-tellmewhy}}. 
TellMeWhy dataset comprises free-form why-questions related to events in short sections. The dataset was created using template-based transformations to generate questions, with crowdsourcing to gather answers. Sections were sourced from ROCStories \citep{mostafazadeh-etal-2016-corpus}, a similar domain to the training dataset (FairytaleQA). TellMeWhy contains a mixture of explicit and implicit questions. Approximately 28.82\% of questions in the dataset are implicit. We evaluate with 1,134 sections and 10,689 questions from the test split.
\\
\\
\textbf{SQuAD1.1 \citep{rajpurkar-etal-2016-squad}}.
Squad1.1 dataset is a comprehensive benchmark that focuses on machine comprehension, question generation, and question answering tasks. It consists of a large collection of articles from Wikipedia, covering a wide range of topics, which is a different source from the training dataset (FairytaleQA). Each article is accompanied by a set of only explicit questions. We evaluate with 2,429 sections, and 12,010 questions from the SQuAD1.1 test split created by \citet{du-etal-2017-learning}.

\begin{table*}[ht]
\centering
\resizebox{0.55\textheight}{!}{\begin{tabular}{lr r r r r r}
\hline
\toprule
\multicolumn{7}{c}{\begin{tabular}[c]{@{}c@{}c@{}} \textbf{TellMeWhy} \end{tabular}} \\
\midrule
\multirow{1}{*}{\textbf{Architecture}} & \multicolumn{1}{c}{\begin{tabular}[c]{@{}c@{}c@{}} \textbf{\# Generated} \\ \textbf{Questions} \\ \textbf{Per Section} \end{tabular}} & \multicolumn{1}{c}{\begin{tabular}[c]{@{}c@{}c@{}} \textbf{\# Answerable} \\ \textbf{Questions} \\ \textbf{Per Section} $\uparrow$ \\ \end{tabular}} & \multicolumn{1}{c}{\begin{tabular}[c]{@{}c@{}}\textbf{Rouge-L} \\ \textbf{F1} $\uparrow$ \\ \end{tabular}} & \multicolumn{1}{c}{\begin{tabular}[c]{@{}c@{}}\textbf{BERTScore} \\ \textbf{F1} $\uparrow$ \\ \end{tabular}} & \multicolumn{1}{c}{\textbf{BLEURT} $\uparrow$} & \multicolumn{1}{c}{\textbf{Self-BLEU} $\downarrow$} \\
\midrule 
EQG        & 4.00 & 0.63 & 35.91 & 0.9129 & 0.4126 & 0.9425 \\
QAG        & 1.53 & 0.45 & 30.35 & 0.9231 & 0.4360 & - \\
mQG   & 4.00 & \textbf{2.10} & \textbf{56.17} & \textbf{0.9361} & \textbf{0.5475} & \textbf{0.3191} \\
\bottomrule
\hline
\end{tabular}}
\caption{Zero-shot evaluation result on TellMeWhy dataset.
Due to a low number of questions, Self-BLEU which cannot be measured is marked with a hyphen.
mQG shows the highest semantic similarity scores with more diversity and generates the largest number of answerable questions.
}
\label{table:zero_eval1}
\end{table*}

\begin{table*}[ht]
\centering
\resizebox{0.55\textheight}{!}{\begin{tabular}{lr r r r r r}
\hline
\toprule
\multicolumn{7}{c}{\begin{tabular}[c]{@{}c@{}c@{}} \textbf{SQuAD1.1} \end{tabular}} \\
\midrule
\multirow{1}{*}{\textbf{Architecture}} & \multicolumn{1}{c}{\begin{tabular}[c]{@{}c@{}c@{}} \textbf{\# Generated} \\ \textbf{Questions} \\ \textbf{Per Section} \end{tabular}} & \multicolumn{1}{r}{\begin{tabular}[c]{@{}c@{}c@{}} \textbf{\# Answerable} \\ \textbf{Questions} \\ \textbf{Per Section} $\uparrow$ \\ \end{tabular}} & \multicolumn{1}{r}{\begin{tabular}[c]{@{}c@{}}\textbf{Rouge-L} \\ \textbf{F1} $\uparrow$ \\ \end{tabular}} & \multicolumn{1}{c}{\begin{tabular}[c]{@{}c@{}}\textbf{BERTScore} \\ \textbf{F1} $\uparrow$ \\ \end{tabular}} & \multicolumn{1}{c}{\textbf{BLEURT} $\uparrow$} & \multicolumn{1}{r}{\textbf{Self-BLEU} $\downarrow$} \\
\midrule
EQG        & 28.00 & 3.74 & 30.31 & 0.8977 & 0.4219 & 0.9695 \\
QAG        & 29.77 & 14.40 & \textbf{46.75} & 0.9203 & 0.5265 & 0.7172 \\
mQG   & 28.00 & \textbf{20.15} & 45.38 & \textbf{0.9211} & \textbf{0.5508} & \textbf{0.6157} \\
\bottomrule
\hline
\end{tabular}}
\caption{Zero-shot evaluation result on SQuAD1.1 dataset. 
mQG generates the most answerable questions with more diversity.
}
\label{table:zero_eval2}
\end{table*}

\subsubsection{Zero-shot Results}
In zero-shot evaluation, we compare mQG with two strong baselines, EQG and QAG. Initially, we examine the performance on the Tellmewhy dataset in Table \ref{table:zero_eval1}. Given that the TellMeWhy dataset only contains why-questions, we select why-questions from the generated questions for evaluation. mQG achieved the highest semantic similarity scores and outperformed baseline models in terms of the number of answerable questions and exhibited better diversity. Zero-shot evaluation on the Tellmewhy dataset, which contains a mix of explicit and implicit questions, demonstrates the ability of mQG to generate different question styles based on answers effectively.

Table \ref{table:zero_eval2} shows evaluation results on the SQuAD1.1 dataset. Even with an out-of-domain dataset, mQG still demonstrates notable performance. mQG outperforms in generating diverse questions and producing a greater number of answerable questions compared to other baselines. However, in the Rouge-L F1 score, mQG is slightly lower than QAG. This can be attributed to the exclusive focus of the SQuAD dataset on explicit questions, and the answer-aware question generation method used by QAG, which is renowned for its effectiveness in generating explicit questions.  Yet, when employing embedding-based evaluation methods such as BERTScore and BLEURT, mQG outperforms the baseline models, particularly in the case of BLEURT. The fact that mQG still demonstrates decent performance on the SQuAD dataset, despite the limitation of the dataset to explicit questions and its status as an out-of-domain dataset, further emphasizes the effectiveness of mQG.

Through these two different settings, we see promising results of mQG. It shows the adaptability of mQG to diverse question styles and domains, further validating the robustness and utility of mQG.

\section{Ablation Study}
\label{sec:sys}
\begin{figure}[t]
\centering
{\includegraphics[width=\columnwidth]{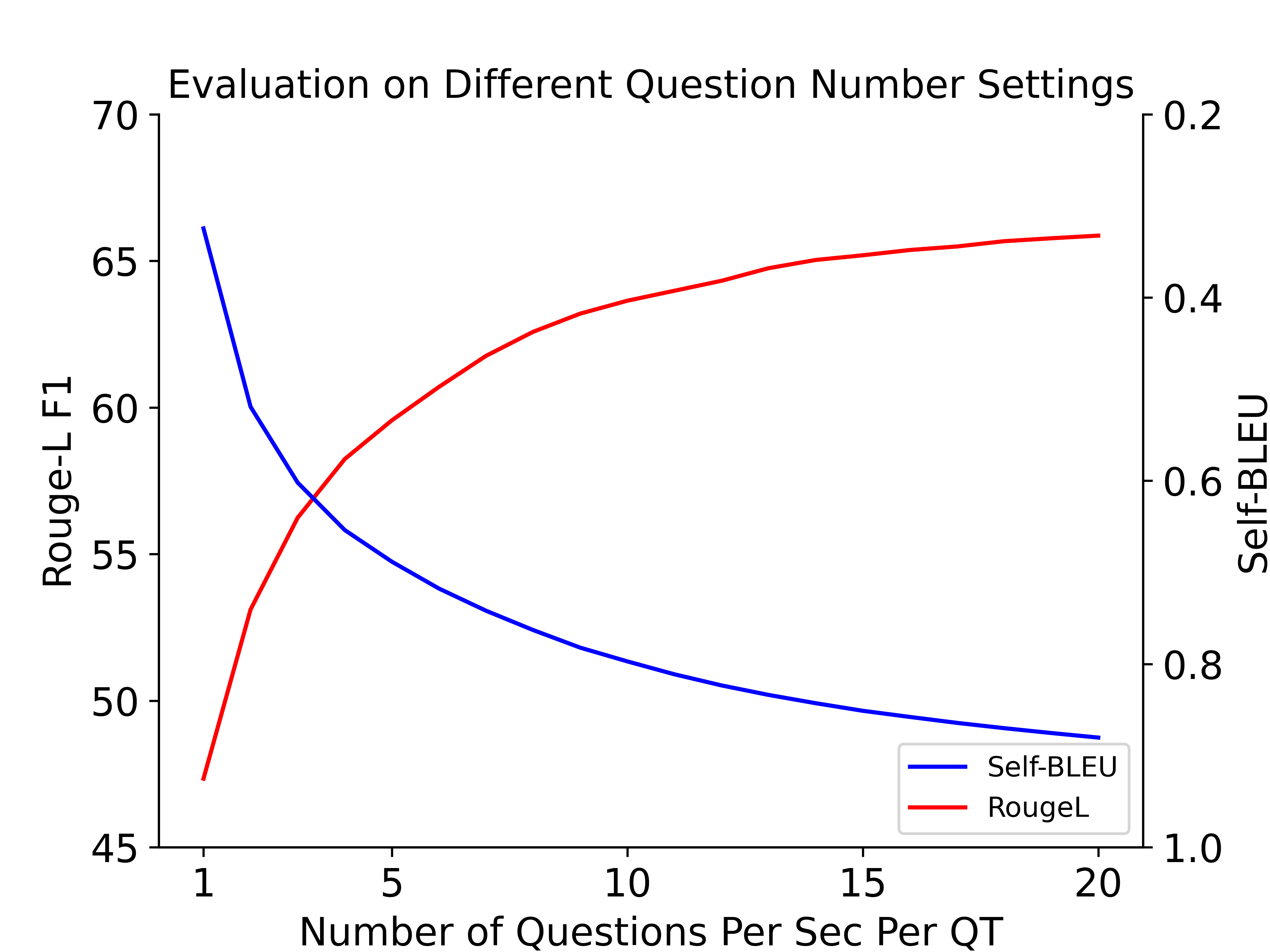}}
\caption{
Results of different question number settings on the original FairytaleQA test set.
Self-BLEU is presented here in a reversed format to allow for a more intuitive visual comparison. 
Intersections of the curves represent the optimal trade-off between two metrics.
}
\label{fig:setting} 
\end{figure}

\begin{table*}[ht]
\centering
\resizebox{0.63\textheight}{!}{\begin{tabular}{lrr rr rr cc cc cc}
\hline
\toprule
\multicolumn{13}{c}{\begin{tabular}[c]{@{}c@{}c@{}} \textbf{FairytaleQA} \\ \end{tabular}} \\
\midrule
\multirow{2}{*}{\textbf{Architecture}} & \multicolumn{2}{c}{\begin{tabular}[c]{@{}c@{}c@{}} \textbf{\# Generated} \\ \textbf{Questions} \\ \textbf{Per Section} \end{tabular}} & \multicolumn{2}{c}{\begin{tabular}[c]{@{}c@{}c@{}} \textbf{\# Answerable} \\ \textbf{Questions} \\ \textbf{Per Section} $\uparrow$ \\ \end{tabular}} & \multicolumn{2}{c}{\begin{tabular}[c]{@{}c@{}}\textbf{Rouge-L} \\ \textbf{F1} $\uparrow$ \\ \end{tabular}} & \multicolumn{2}{c}{\begin{tabular}[c]{@{}c@{}}\textbf{BERTScore} \\ \textbf{F1} $\uparrow$ \\ \end{tabular}} & \multicolumn{2}{c}{\textbf{BLEURT} $\uparrow$} &\multicolumn{2}{c}{\textbf{Self-BLEU} $\downarrow$}  \\ 
\cmidrule(lr){2-3} \cmidrule(lr){4-5} \cmidrule(lr){6-7} \cmidrule(lr){8-9} \cmidrule(lr){10-11} \cmidrule(lr){12-13}
& \multicolumn{1}{c}{M} & \multicolumn{1}{c}{SE} & \multicolumn{1}{c}{M} & \multicolumn{1}{c}{SE} & \multicolumn{1}{c}{M} & \multicolumn{1}{c}{SE} & \multicolumn{1}{c}{M} & \multicolumn{1}{c}{SE} & \multicolumn{1}{c}{M} & \multicolumn{1}{c}{SE} & \multicolumn{1}{c}{M} & \multicolumn{1}{c}{SE} \\
\midrule
mQG               & 28.00 & 0.00 & \textbf{23.08} & 0.36 & \textbf{58.90} & 0.37 & \textbf{0.9394} & 0.0005 & 0.5698 & 0.0033 & \textbf{0.6389} & 0.0079 \\
- $ \mathcal{L}_{MQS}$ & 28.00 & 0.00 & 22.67 & 0.28 & 58.66 & 0.08 & \textbf{0.9394} & 0.0003 & \textbf{0.5703} & 0.0019 & 0.7006 & 0.0045 \\
- $ \mathcal{L}_{MQS} $ \& reference questions & 28.00 & 0.00 & 22.65 & 0.41& 54.76 & 0.22 & 0.9353 & 0.0005 & 0.5428 & 0.0011 & 0.7529 & 0.0032\\
\bottomrule
\hline
\end{tabular}}
\caption{
The comparison results of mQG with and without maximum question similarity loss and reference questions.
}
\label{ablation}
\end{table*}

\subsection{Setting of Question Number}

Given that mQG can be set with the number of questions to generate, we conduct an experiment on various settings of question number per section per question type to generate. In Figure \ref{fig:setting}, the evaluation result is based on the original FairytaleQA test set. As the quantity of generated questions increases,  the Rouge-L F1 score provides satisfactory results, though diversity decreases. This indicates that a significant increase in the number of generated questions tends to produce similar questions with different phrasings. Setting the number of generated questions at 4 shows the optimal trade-off between the Rouge-L F1 and the Self-BLEU.

\subsection{Analysis of Maximum Question Similarity Loss and Recursive Framework}
\label{sec:qcl}
As discussed in section \ref{sec:qcl}, mQG aims to increase diversity within questions while maintaining semantic correctness. mQG w/o $\mathcal{L}_{MQS}$ refers to the mQG model only trained with $\mathcal{L}_{CE}$. For mQG w/o $\mathcal{L}_{MQS}$ and reference questions, we give only question type and context as input while training, and no recursive framework is used in inference. Table \ref{ablation} shows that the mQG model with maximum question similarity loss $\mathcal{L}_{MQS}$ and reference questions hugely increase diversity. Additionally, the number of answerable questions has also improved. This could be attributed to the fact that all ground-truth questions are answerable, and mQG maximizes the similarity between these questions and continually references the most probable question during inference. These results indicate that each framework of mQG effectively enhances the probability of generating a diverse set of possible questions.

\section{Conclusion}


In this work, we extend the scope of answer-unaware question generation to generate multiple diverse questions. We propose a novel framework that applies a maximum question similarity loss during training to promote question diversity, followed by a recursive generation process for further refinement. Additionally, an evaluation model is introduced to verify the answerability of the generated questions. Recognizing the essential role of narrative questions in education, we train and evaluate mQG accordingly. Comprehensive experiments validate the efficacy of mQG across a variety of datasets, highlighting its potential utility in environments that demand diverse narrative questions.

\section*{Limitations}
\label{sec:limitation}

mQG framework utilizes a recursive feedback mechanism for generating questions during the inference stage. However, the quality of these generated questions remains uncertain. If the quality of previously generated questions is poor, this may adversely impact the quality of subsequent questions produced by mQG. Moreover, the quantity of questions that can be generated is limited by a maximum token threshold. Another limitation is the potential risk of misclassification by the evaluation model, which could lead to the categorization of unanswerable questions as answerable. Despite our efforts to mitigate this risk, the evaluation model is still at a level of uncertainty in accurately classifying the generated questions. Even with the fact that reliability scores can be low in NLP tasks, in the quality human evaluation, the reliability scores are relatively low. This can lead to uncertainty in the results.

\section*{Ethics Statement}
\label{sec:ethics}
The results are appropriately placed in the context of prior and existing research. All generation models are trained on the FairytaleQA dataset which is publicly available and has no ethical issues as annotated by educational experts. In the human evaluation process, we pay annotators more than the minimum wage.

\section*{Acknowledgements}
We would like to thank the anonymous reviewers for their helpful questions and comments.
JinYeong Bak is the corresponding author.
This work was partly supported by Institute of Information \& communications Technology Planning \& Evaluation (IITP) grant funded by the Korea government (MSIT) (No.2022-0-00680, Abductive inference framework using omni-data for understanding complex causal relations \& No.2019-0-00421, AI Graduate School Support Program (Sungkyunkwan University)), and a grant from the National Research Foundation of Korea (NRF) [NRF-2021R1A4A3033128].

\bibliography{anthology, custom}
\bibliographystyle{acl_natbib}

\section*{Appendix}
\appendix

\section{Further Analysis on Evaluation Model}
\label{sec:fa_evalationmodel}

\subsection{Preprocessing Dataset}
To evaluate each cross-validation set with an answerability evaluation model, we train the evaluation model with different FairytaleQA trainsets. One is an originally constructed trainset and the others are randomly split by books. From the FairytaleQA dataset, some explicit questions were not able to be found in the section and some questions with cross-annotated answers had different aspects of answers (explicit, implicit). We removed those questions and a number of total questions after preprocessing is described in Table \ref{table:stat_em}.

\begin{figure}[t!]
\centering
{\includegraphics[scale=0.9]{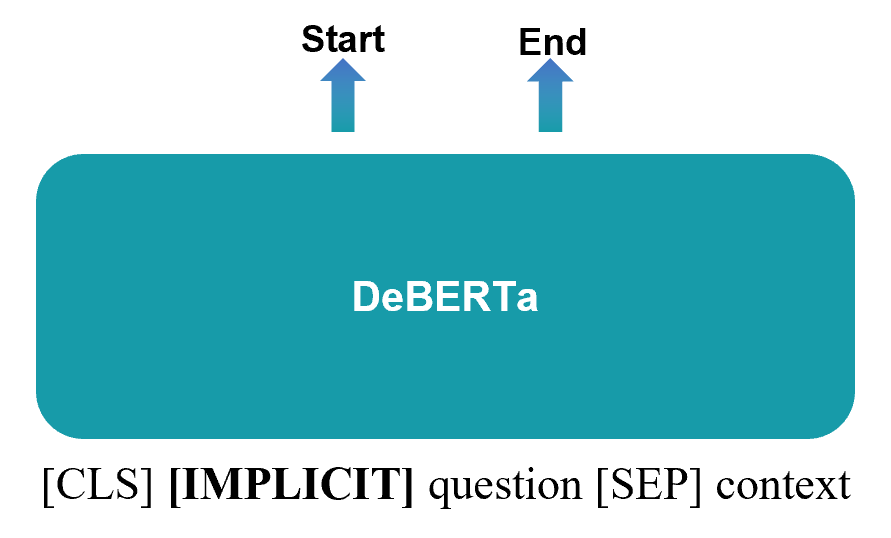}}
\caption{Overview of Answerability Evaluation Model.}
\label{fig:aem} 
\end{figure}

\begin{table}[t!]
\centering
\resizebox{0.8\columnwidth}{!}{\begin{tabular}{lccc}
\hline
\toprule
 & \multicolumn{1}{l}{Explicit} & \multicolumn{1}{l}{Implicit} & \multicolumn{1}{l}{Total} \\
 \midrule
\# questions & 5,376 & 1,963 & 7,309 \\
\bottomrule
\hline
\end{tabular}}
\caption{
The number of questions of the FairytaleQA dataset after annotation mistakes were removed.
}
\label{table:stat_em}
\end{table}

\begin{figure}[t!]
\centering
{\includegraphics[width=\columnwidth]{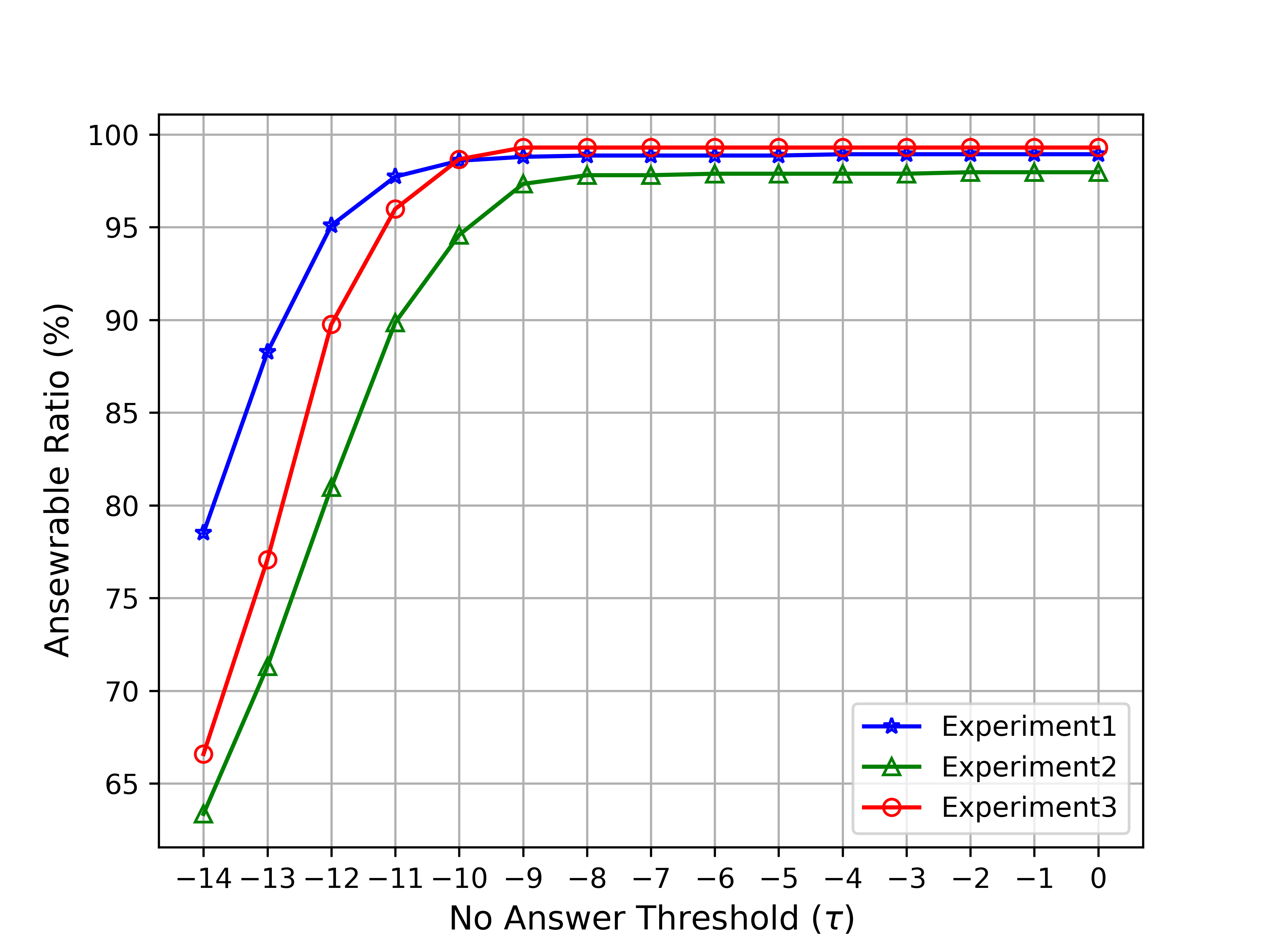}}
\caption{
The answerable ratio of val+test set by different threshold settings.
}
\label{fig:ansratio} 
\end{figure}

\subsection{Evaluation Model Postprocessing}
\label{sec:EMT}

In terms of post-processing, we take a similar approach by \citet{devlin-etal-2019-bert}. Classified results $y_c$ of each question are formulated as:
\begin{equation}
\label{postprocess}
\resizebox{\columnwidth}{!}{\begin{math}\begin{aligned}
  y_{c} =
    \begin{cases}
    \multirow{2}{*}{\text{No Answer,}} & \text{if } CLS_{se}  > a_{se} + \tau \\ & \text{and }  CLS_{se}  > IMP_{se} + \tau \\
      \text{Implicit,} & \text{else if } IMP_{se} > a_{se}  \\
      \text{Explicit,} & \text{otherwise}. \\
    \end{cases}  
\end{aligned}\end{math}}
\end{equation}
$CLS_{se}$ denotes score of \texttt{[CLS]} token as answer start span and answer end span. $IMP_{se}$ denotes score of \texttt{[IMP]} token as answer start span and answer end span. $a_{se}$ denotes the best score of answer start span and answer end span without \texttt{[CLS]} and \texttt{[IMP]}. Additionally, if an answer end span indice is lower than an answer start span indice we classify it as no answer. Threshold $\tau$ is selected on the ground-truth set to maximize the performance. This threshold is set differently for each evaluation model. Figure \ref{fig:ansratio} shows the answerable ratio percentage by different threshold settings. We also train three evaluation models with each train set for cross-validation in the main results. We select each threshold before a significant drop in the answerable ratio is observed. -12, -10, and -11 are each threshold for experiment1, experiment2, and experiment3.

\begin{table}[t!]
\centering
\resizebox{0.75\columnwidth}{!}{\begin{tabular}{lrr rr}
\hline
\toprule
& \multicolumn{2}{c}{\textbf{F1}} & \multicolumn{2}{c}{\textbf{Accuracy}} \\ 
\cmidrule(lr){2-3} \cmidrule(lr){4-5}
& \multicolumn{1}{c}{M} & \multicolumn{1}{c}{SE} & \multicolumn{1}{c}{M} & \multicolumn{1}{c}{SE}\\
\midrule
Explicit & 78.72 & 0.52 & 88.26 & 0.17 \\
Implicit & 64.76 & 2.05 & 64.76 & 2.05 \\
Total & 75.28 & 1.01 & 82.49 & 0.81 \\
\bottomrule
\hline
\end{tabular}}
\caption{
Ground-truth val+test set results on the evaluation model. Each model is trained with each cross-validation trainset. 
}
\label{table:res_EM}
\end{table}

\subsection{Evaluation Model Results}

We perform cross-validation to measure the performance of the main results in Table \ref{table:main}, and as a result, we train each evaluation model with each trainset. Since our goal is to classify questions as explicit, implicit, or unanswerable, we count explicit questions as accurate if at least one of the predicted answer tokens is found in the ground-truth answer. This is denoted as "Accuracy" in Table \ref{table:res_EM}. The F1 measurement follows the implementation by \citet{devlin-etal-2019-bert}. The evaluation model classifies explicit questions more accurately than implicit questions.

\subsection{Classified Questions Analysis}

\begin{table}[t!]
\centering
\resizebox{\columnwidth}{!}{\begin{tabular}{lr r r r r}
\hline
\toprule
\multicolumn{6}{c}{\begin{tabular}[c]{@{}c@{}c@{}} \textbf{FairytaleQA} \\ \end{tabular}} \\
\midrule
& \multicolumn{1}{l}{\textbf{Ground-truth}} & \multicolumn{1}{l}{\textbf{QAG (top10)}} & \multicolumn{1}{l}{\textbf{QAG}} & \multicolumn{1}{l}{\textbf{EQG}} &  \multicolumn{1}{l}{\textbf{mQG}} \\ \hline
Explicit & 74.10\% & 79.05\% & 71.69\% & 54.42\% & 60.65\% \\
Implicit & 21.22\% & 2.50\% & 5.08\% & 33.95\% & 22.38\% \\
No Ans. & 4.68\% & 18.45\% & 23.23\% & 11.63\% & 16.97\% \\
\midrule
Total & 919 & 2,835 & 7,534 & 1,402 & 8,820\\
\bottomrule
\hline
\end{tabular}}
\caption{
The FairyTaleQA test set analysis of questions by answer types, classified by evaluation model.  
Total denotes the number of questions after duplicates from the same context are removed.
Each answer type is denoted with a proportion in each model.
}
\label{table:ans}
\end{table}

We analyze the ratio of questions classified into different answer types by the answerability evaluation model. Even though the ground-truth questions do not contain unanswerable questions, the evaluation model classifies approximately 4.5\% of the questions as unanswerable, as shown in Table 5. The problem of answer-aware question generation is well-known. QAG uses the answer as an input in the question generation process, and our results show that QAG is not fit for generating implicit questions, as only about 5.1\% of questions are classified as implicit. The EQG baseline generates both explicit and implicit questions but only has a small number of total questions after removing duplicates. On the other hand, the mQG still has a large number of questions even after removing duplicates, totaling 8,820, with explicit and implicit questions roughly in a 3-to-1 ratio. These results show that the mQG generates both types of multiple questions better than other baselines.

\section{Diversity Exploration}

\begin{table}[t!]
\centering
\resizebox{\columnwidth}{!}{\begin{tabular}{ll}
\hline
\toprule
\multicolumn{1}{r}{\textbf{Self-BLEU}} & \multicolumn{1}{c}{\textbf{Example Questions}}\\
\midrule
\multirow{4}{*}{0.3150} & Why did the Dragon King want to capture a monkey? \\
& Why couldn't the Dragon King's servants capture a monkey? \\
& Why did the Dragon King consult his chief steward? \\
& Why was the Dragon King greatly puzzled? \\
\midrule
\multirow{4}{*}{0.6362} & Why did the Dragon King want to capture a monkey? \\
& Why couldn't the Dragon King's servants capture a monkey? \\
& Why did the Dragon King consult his chief steward? \\
& How did the Dragon King consult his chief steward? \\
\midrule
\multirow{4}{*}{0.7830} & Why did the Dragon King consult his chief? \\
& Why did the Dragon King consult steward? \\
& Why did the Dragon King consult his chief steward? \\
& How did the King consult his chief steward? \\
\midrule
\multirow{4}{*}{0.9014} & Why did the Dragon King consult his chief steward? \\
& Why did the Dragon King consult his chief? \\
& Why did the Dragon King consult his chief steward? \\
& How did the Dragon King consult his chief steward? \\
\bottomrule
\end{tabular}}
\caption{Examples on Self-BLEU scores with 4 questions each.}
\label{tab:div_examples}
\end{table}

For diversity evaluation, we calculate the Self-BLEU score among generated questions from the same context. Self-BLEU score is based on BLEU evaluation method \citep{papineni-etal-2002-bleu}. The BLEU evaluation method has many criticisms for evaluating sentence-level corpus. If a higher-order n-gram precision goes to 0, the total BLEU score goes to 0. As an outcome, many variations applying the smoothing method for the BLEU score have shown \citep{chen-cherry-2014-systematic}. We apply 'smoothing 1' described in \citet{chen-cherry-2014-systematic} since all the generated questions are sentence-level. Examples of Self-BLEU scores are shown in table \ref{tab:div_examples}. When the Self-BLEU score goes up to 0.7830, almost all questions can be addressed by the same answers.

\section{Decoding Method and Model Selection}

\begin{table}[t!]
\centering
\resizebox{\columnwidth}{!}{\begin{tabular}{l c r r r}
\hline
\toprule
\multicolumn{5}{c}{\begin{tabular}[c]{@{}c@{}c@{}} \textbf{FairytaleQA} \\ \end{tabular}} \\
\midrule
\multirow{1}{*}{\textbf{Architecture}} & \multirow{1}{*}{\textbf{Decdoing Method}} & \multicolumn{1}{c}{\begin{tabular}[c]{@{}c@{}c@{}} \textbf{\# Answerable} \\ \textbf{Questions} \\ \textbf{Per Section} $\uparrow$ \\ \end{tabular}} & \multicolumn{1}{c}{\begin{tabular}[c]{@{}c@{}}\textbf{Rouge-L} \\ \textbf{F1} $\uparrow$ \\ \end{tabular}} & \multicolumn{1}{c}{\textbf{Self-BLEU} $\downarrow$} \\
\midrule
mQG-T5                 & b=5 & 17.89 &	30.59 &	\textbf{0.5476} \\
mQG-BART               & b=5 & \textbf{23.35} & \textbf{58.24} & 0.6243 \\
                       & p=0.1 & 16.89  &	53.45 &	0.7826 \\
                       & p=0.5 & 18.01  &	53.54 &	0.7622 \\
                       & p=0.75 & 19.12 &	54.45 &	0.7321 \\
                       & p=0.95 & 20.06 &	54.90 &	0.7135 \\
\bottomrule
\hline
\end{tabular}}
\caption{
Performance of mQG with different backbone models and decoding methods on the original test set. 
b=5 denotes beam search with beam size set to 5. 
p denotes nucleus sampling (NS@p; $p \in {0.1, 0.5, 0.75, 0.95}$). 
All models are set to generate 28 questions per section.
}
\label{tab:model_select}
\end{table}

Moreover, in addition to the main results, we compare the performance of mQG between different backbone models and decoding methods. In Table \ref{tab:model_select}, T5-based mQG exhibits the best Self-BLEU score but significantly lags behind BART-based mQG in terms of \# Answerable Questions Per Section and Rouge-L score. This suggests that T5-based mQG struggles to generate semantically correct questions. When comparing decoding methods, beam search outperforms nucleus sampling in all dimensions. This is due to the decoding process of mQG, which returns multiple sequences to exclude pre-generated questions. Beam search utilizes a tree search algorithm, whereas nucleus sampling does not. As a result, nucleus sampling tends to generate duplicate questions.

\section{Weighting Factor Impact on Performance}

\begin{table}[t!]
\centering
\resizebox{0.8\columnwidth}{!}{\begin{tabular}{l r r r}
\hline
\toprule
\multicolumn{4}{c}{\begin{tabular}[c]{@{}c@{}c@{}} \textbf{FairytaleQA} \\ \end{tabular}} \\
\midrule
\multirow{1}{*}{\textbf{$\beta$}} & \multicolumn{1}{c}{\begin{tabular}[c]{@{}c@{}c@{}} \textbf{\# Answerable} \\ \textbf{Questions} \\ \textbf{Per Section} $\uparrow$ \\ \end{tabular}} & \multicolumn{1}{c}{\begin{tabular}[c]{@{}c@{}}\textbf{Rouge-L} \\ \textbf{F1} $\uparrow$ \\ \end{tabular}} & \multicolumn{1}{c}{\textbf{Self-BLEU} $\downarrow$} \\
\midrule
0.0 & 22.89 &	58.49 & 	0.4747 \\
0.2 & 23.16 &	59.40 & 	0.4117 \\
0.4 & 23.23 &	\textbf{59.54} &     \textbf{0.4052} \\
0.6 & 23.26 &	58.44 & 	0.4261 \\
0.8 & 23.34 &	59.29 & 	0.4288 \\
1.0 & 23.35 &	58.24 & 	0.4210 \\
2.0 & 23.28 &	58.28 & 	0.4297 \\
3.0 & 23.34 &	58.42 & 	0.4478 \\
5.0 & \textbf{23.50} &	58.15 & 	0.4527 \\
\bottomrule
\hline
\end{tabular}}
\caption{
mQG results on different $\beta$ settings on the original test set.
0.0 equals to mQG w/o maximum question similarity loss $\mathcal{L}_{MQS}$.
All models are set to generate 28 questions per section.
}
\label{tab:weight}
\end{table}

\begin{table}[t!]
\centering
\resizebox{\columnwidth}{!}{\begin{tabular}{l r r r}
\hline
\toprule
\multirow{1}{*}{\textbf{Architecture}} & \multicolumn{1}{c}{\textbf{Rouge-L (ori)}} & \multicolumn{1}{c}{\textbf{Rouge-L (alt)}} & \multicolumn{1}{c}{\textbf{Diff}} \\
\midrule
\multicolumn{4}{c}{\textbf{FairytaleQA}} \\
\midrule
EQG           & 41.05 & 39.35 & 1.70\\
QAG           & 53.77 & 53.13 & 0.64\\
mQG           & 58.90 & 58.36 & 0.54\\
\midrule
\multicolumn{4}{c}{\textbf{TellMeWhy}} \\
\midrule
EQG           & 35.91 & 15.08 & 20.83 \\
QAG           & 30.35 & 23.93 & 6.42 \\
mQG           & 56.17 & 51.57 & 4.60 \\
\midrule
\multicolumn{4}{c}{\textbf{SQuAD1.1}} \\
\midrule
EQG           & 30.31 & 25.84 & 4.47 \\
QAG           & 46.75 & 44.85 & 1.90 \\
mQG           & 45.38 & 43.20 & 2.18 \\
\bottomrule
\hline
\end{tabular}}
\caption{
Comparison results on Rouge-L calculation.
FairytaleQA results are the mean value of 3 cross-validation results.
Rouge-L (alt) denotes one-to-one match calculation.
Diff denotes the difference between Rouge-L (ori) and Rouge-L (alt).
}
\label{tab:Rouge-L}
\end{table}

To determine how MQS loss affects training, we conduct experiments with the mQG model using different settings for the weighting factor $\beta$. The overall training objective $\mathcal{L}$ is defined as 
\begin{equation}
\begin{aligned}
\mathcal{L} = \mathcal{L}_{CE} + \beta * \mathcal{L}_{MQS} 
\end{aligned}
\end{equation}
In Table \ref{tab:weight}, Self-BLEU is calculated between questions that share context and question type. The optimal point of diversity is achieved when $\beta$ is set to 0.4. As $\beta$ increases, the Self-BLEU score decreases, while the number of answerable questions increases. This outcome aligns with our goal of implementing MQS loss to enhance diversity within the bounds of semantic correctness.

\section{Another Rouge-L Calculation}
As mentioned in Section \ref{sec:autoeval}, we calculate the Rouge-L score only to find the highest score for each ground-truth question. This calculation method may lead to the one-to-many matching problem. To determine if the problem has occurred, we compare the results with another Rouge-L calculation Rouge-L (alt). This calculation excludes previously matched generated questions, allowing for only one-to-one matches. In Table \ref{tab:Rouge-L}, most Rouge-L (alt) results exhibit slightly lower scores in comparison to Rouge-L (ori), suggesting that one-to-many problems have occurred, although the impact is relatively minor as the ground-truth questions are a unique set of questions. The significant difference in the TellMeWhy dataset can be attributed to the limited number of 'why' questions generated.

\section{Implementation Details}
For the mQG model, we use the MQS loss of the validation set as the selecting criteria. For the mQG models without MQS loss, we use MLE loss as the selecting criteria. Total training time was about 3 hours with 1 RTX A6000 GPU. We initialize the mQG model with pretrained BART-large, which has 406M parameters. Hyperparameters are follow: learning rate = 5e-6; batch size = 8; epoch = 15
\\
\\
We use RoBERTa-large model for BERTScore and BLEURT-20 model for BLEURT.
For the evaluation model, we load SQuAD 2.0 finetuned DeBERTa-base model \footnote{\url{https://huggingface.co/deepset/deberta-v3-base-squad2}}, which has 86M parameters, to further finetune. Total training time was about an hour with 1 RTX A6000 GPU. Hyperparameters are follow: learning rate = 5e-6; batch size = 16; epoch = 8


\section{Examples of Generated Questions}
Tables \ref{tab:ex2} and \ref{tab:ex1} show the generated examples of the mQG, EQG, QAG, and ground truth questions with the according section and classified results with the answerability evaluation model. Even with different settings for generating multiple questions, EQG still generated duplicate questions because it guided the model only with special tokens to generate multiple questions. QAG has generated different questions but with less diversity. In all questions, the evaluation model accurately classified the questions. Given the sufficient number of questions generated by each model, we selected four questions as representative examples. Given the sufficient number of questions generated by each model, we selected 4 questions as representative examples.

\begin{table*}[!htbp]
\begin{tabularx}{\textwidth}{X}
\hline
\toprule
\textbf{Section} \\
But his brother complained of being weary, and at length they decided to remain there for the night. When Andrew awoke he found himself alone; and he saw neither brother nor boat, until he came to the highest point of the island. Then he discovered him far out, darting for land like a sea-gull. Andrew did not understand the whole affair. There were still provisions there, as well as a dish of curd, his gun and various other things. So Andrew wasted but little time in thought. "He will come back this evening," said he. "Only a fool loses heart so long as he can eat." But in the evening there was no brother to be seen, and Andrew waited day by day, and week by week; until at last, he realized that his brother had marooned him on this barren island in order to be able to keep their inheritance for himself, and not have to divide it. And such was the case, for when John Nicholas came in sight of land on his homeward trip, he had capsized the boat, and declared that Lucky Andrew had been drowned.
\\ \hline
\textbf{Ground-truth Questions} \\
What was John Nicholas doing when Andrew saw him? (Explicit) \\
Why did John Nicholas capsize the boat when he reached land? (Implicit) \\
Why did Andrew want the inheritance to himself? (Implicit) \\
How did Andrew feel when he saw his brother and boat far out? (Implicit)
\\ \hline
\textbf{mQG} \\
What did John Nicholas declare when he came in sight of land? (Explicit) \\
Why did John Nicholas marooned his brother on a barren island? (Explicit) \\
Why did the brother want to keep their inheritance for himself? (Implicit) \\
Why did John Nicholas declare that Lucky Andrew had been drowned? (Implicit)
\\ \hline
\textbf{QAG} \\
what did andrew find on the island? (Explicit) \\
what did andrew find on the island? (Explicit) \\
who did andrew see after he woke up? (No Answer) \\
when andrew awoke he found himself alone? (No Answer)
\\ \hline
\textbf{EQG} \\
What happened after John Nicholas came in sight of land on his homeward trip? (Explicit) \\
What happened after John Nicholas came in sight of land on his homeward trip? (Explicit) \\
Why did the brother complain of being weary? (Implicit) \\
Why did the brother complain of being weary? (Implicit)
\\
\bottomrule
\hline
\end{tabularx}%
\caption{Example of generated questions of mQG, EQG, and QAG for a given context. On the right of each question, classified results by the answerability evaluation model are shown in parentheses. Generated questions of mQG have the most diverse syntax questions considering question types and still, semantically correspond to the given section.}
\label{tab:ex2}
\end{table*}%

\begin{table*}[!htbp]
\begin{tabularx}{\textwidth}{X}
\hline
\toprule
\textbf{Section} \\
At any rate, his poor young wife was sorely to be pitied, for she was left with a little baby-boy to bring up, and, as times were bad, she had not much to live on.  But she loved her baby dearly, and worked all day amongst her cows, and pigs, and hens, in order to earn enough money to buy food and clothes for both herself and him.  Now, on the morning of which I am speaking, she rose very early and went out to feed her pigs, for rent-day was coming on, and she intended to take one of them, a great, big, fat creature, to the market that very day, as she thought that the price that it would fetch would go a long way towards paying her rent.  And because she thought so, her heart was light, and she hummed a little song to herself as she crossed the yard with her bucket on one arm and her baby-boy on the other.
\\ \hline
\textbf{Ground-truth Questions} \\
What did the widowed Mistress do to earn enough money? (Explicit) \\
Why did the widowed Mistress want to sell her pig? (Explicit) \\
How did the widowed Mistress feel while she thought of selling the pig? (Implicit)
\\ \hline
\textbf{mQG} \\
What did the young woman do because rent-day was coming on? (Explicit) \\
Why did the young wife want to buy a great, big, fat creature? (Explicit) \\
Where did the young woman intend to take one of the pigs? (Explicit) \\
How did the young wife feel to bring up her baby? (Implicit)
\\ \hline
\textbf{EQG} \\
Why did the wife go out to feed her pigs? (Explicit) \\
Why did the wife go out to feed her pigs? (Explicit) \\
How did the wife feel when she saw the pig? (Implicit) \\
How did the wife feel when she saw the pig? (Implicit)
\\ \hline
\textbf{QAG} \\
how did the wife earn money? (Explicit)\\
what day was it? (Implicit) \\
what did the wife feed her pigs? (No Answer) \\
how many pigs did the wife plan to buy? (No Answer)
\\
\bottomrule
\hline
\end{tabularx}%
\caption{Another example of generated questions of mQG, EQG, and QAG for a given context. On the right of each question, classified results by the answerability evaluation model are shown in parentheses. mQG model has generated questions most answerable questions with diversity.}
\label{tab:ex1}%
\end{table*}%

\clearpage
\newpage

\begin{figure}[]
\centering
{\includegraphics[width=\columnwidth]{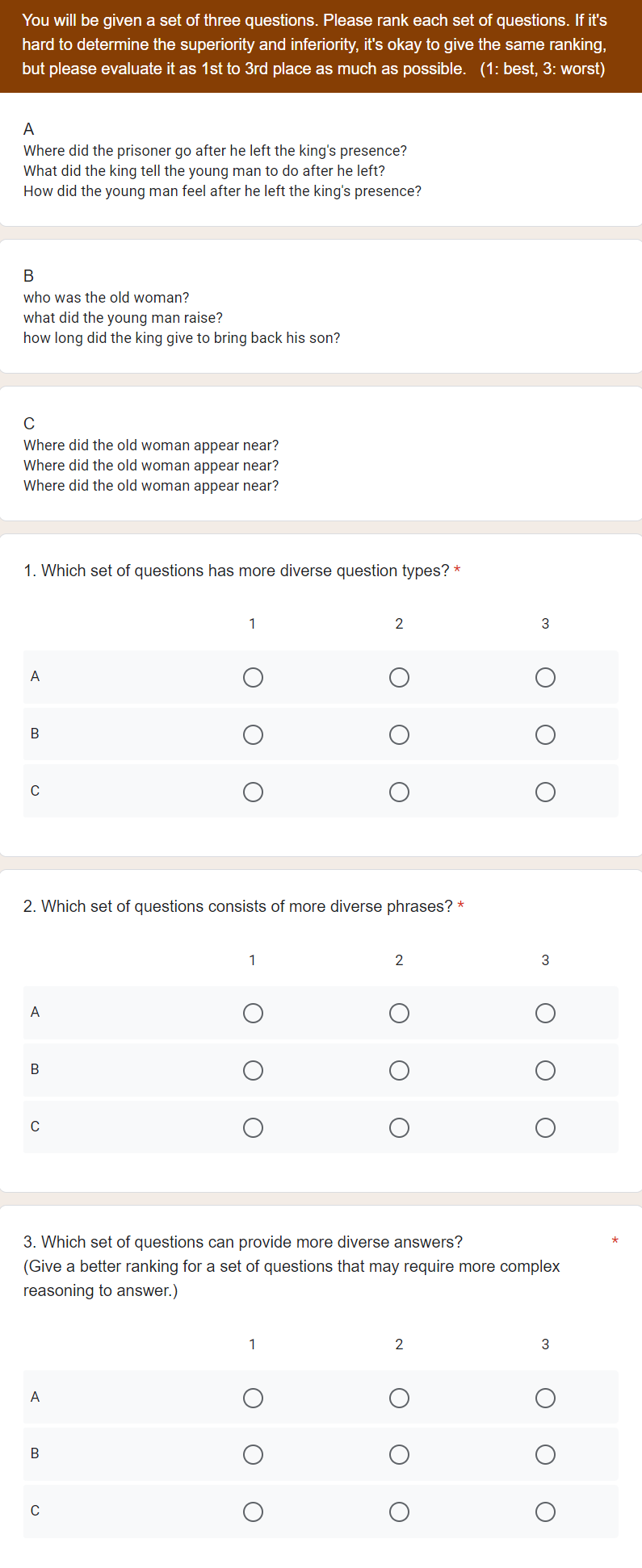}}
\caption{
The question sheet for diversity human evaluation.
}
\label{fig:div_he} 
\end{figure}

\begin{figure}[]
\centering
{\includegraphics[width=\columnwidth]{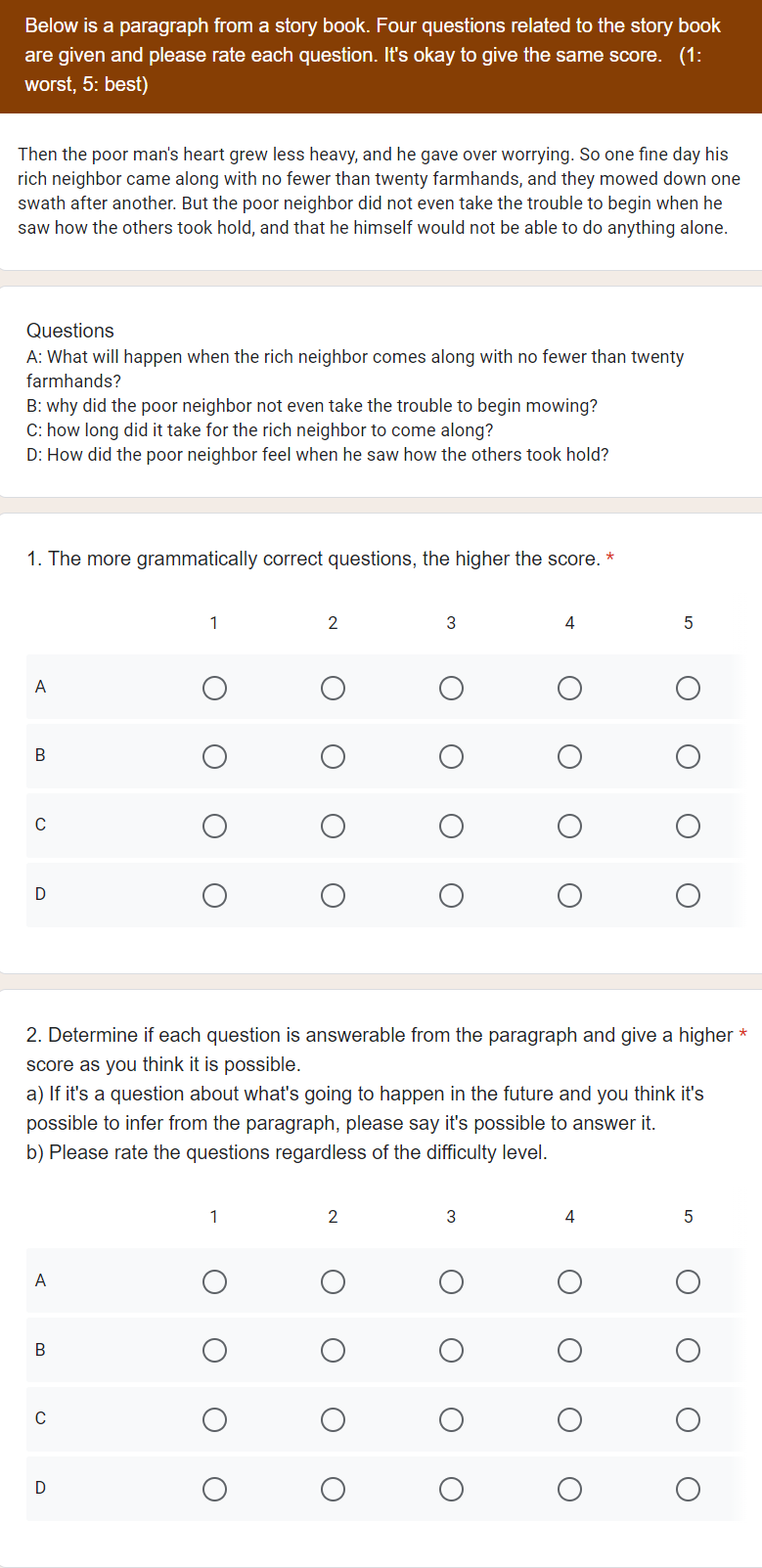}}
\caption{
The question sheet for quality human evaluation.
}
\label{fig:qual_he} 
\end{figure}

\end{document}